\documentclass[runningheads]{llncs}
\usepackage{graphicx}

\usepackage{tikz}
\usepackage{comment}
\usepackage{amsmath,amssymb} 
\usepackage{color}

\usepackage[accsupp]{axessibility}  

\usepackage{textcomp}       
\usepackage{nicefrac}       
\usepackage{url}            
\usepackage{hyperref}       
\usepackage{mycolors}       
\usepackage{xspace}         
\usepackage{mathtools}      
\usepackage{multirow}       
\usepackage{pifont}         
\usepackage{wrapfig}

\begin{document}
\pagestyle{headings}
\mainmatter
\def\ECCVSubNumber{2610}  

\title{A Real World Dataset for Multi-view 3D Reconstruction} 

\titlerunning{A Real World Dataset for Multi-view 3D Reconstruction}
%
\author{
    \small
    Rakesh Shrestha$^{1}$, Siqi Hu$^{2}$, Minghao Gou$^{2, 3}$, Ziyuan Liu$^{2}$, Ping Tan$^{1,2}$ \\
    \normalsize{Simon Fraser University$^{1}$},
    \normalsize{Alibaba XR Lab$^{2}$} \\
    \normalsize{Shanghai Jiao Tong University$^{3}$} \\
    \scriptsize{\{rakeshs,pingtan\}@sfu.ca, \{husiqixiang, ziyuan-liu\}@outlook.com,  gmh2015@sjtu.edu.cn}
    \institute{}
}
\authorrunning{R. Shrestha et al.}
\maketitle

\setcounter{secnumdepth}{3}

\makeatletter
\DeclareRobustCommand\onedot{\futurelet\@let@token\@onedot}
\def\@onedot{\ifx\@let@token.\else.\null\fi\xspace}

\def\eg{\emph{e.g}\onedot} \def\Eg{\emph{E.g}\onedot}
\def\ie{\emph{i.e}\onedot} \def\Ie{\emph{I.e}\onedot}
\def\cf{\emph{c.f}\onedot} \def\Cf{\emph{C.f}\onedot}
\def\etc{\emph{etc}\onedot} \def\vs{\emph{vs}\onedot}
\def\wrt{w.r.t\onedot} \def\dof{d.o.f\onedot}
\def\etal{\emph{et al}\onedot}
\makeatother

\DeclarePairedDelimiter{\norm}{\lVert}{\rVert}

\newcommand{\rakesh}[1]{{\color{Green} \bf \em   [Rakesh: #1]}}
\newcommand{\ping}[1]{{\color{MidnightBlue} \bf \em   [Ping: #1]}}
\newcommand{\ziyuan}[1]{{\color{YellowOrange} \bf \em   [Ziyuan: #1]}}
\newcommand{\siqi}[1]{{\color{Blue} \bf \em   [Siqi: #1]}}
\newcommand{\todo}[1]{{\textcolor{red}{\bf [#1]}}}

\newcommand{\figref}[1]{Figure~\ref{fig:#1}}
\newcommand{\tabref}[1]{Table~\ref{tab:#1}}
\newcommand{\equref}[1]{Equation~(\ref{equ:#1})}
\newcommand{\secref}[1]{Section~\ref{sec:#1}}
\newcommand{\subsecref}[1]{Sub-section~\ref{subsec:#1}}
\newcommand{\tableref}[1]{Table~\ref{table:#1}}
\newcommand{\rows}[1]{\multirow{2}{*}{#1}}

\newcommand{\cmark}{\ding{51}}
\newcommand{\xmark}{\ding{55}}
\newcommand{\starmark}{\ding{81}}


\graphicspath{{./figures/}}

\begin{abstract}
We present a dataset of 998 3D models of everyday tabletop objects along with their 847,000 real world RGB and depth images.
Accurate annotation of camera pose and object pose for each image is performed in a semi-automated fashion
to facilitate the use of the dataset in a myriad 3D applications like shape reconstruction, object pose estimation, shape retrieval \etc.
We primarily focus on learned multi-view 3D reconstruction due to the lack of appropriate real world benchmark for the task
and demonstrate that our dataset can fill that gap.
The entire annotated dataset along with the source code for the annotation tools and evaluation baselines is available at \url{http://www.ocrtoc.org/3d-reconstruction.html}.

\keywords{Dataset, Multi-view 3D reconstruction}
\end{abstract}

\section{Introduction}

\begin{figure*}[t]
	\centering
	\scriptsize
	\includegraphics[width=\textwidth]{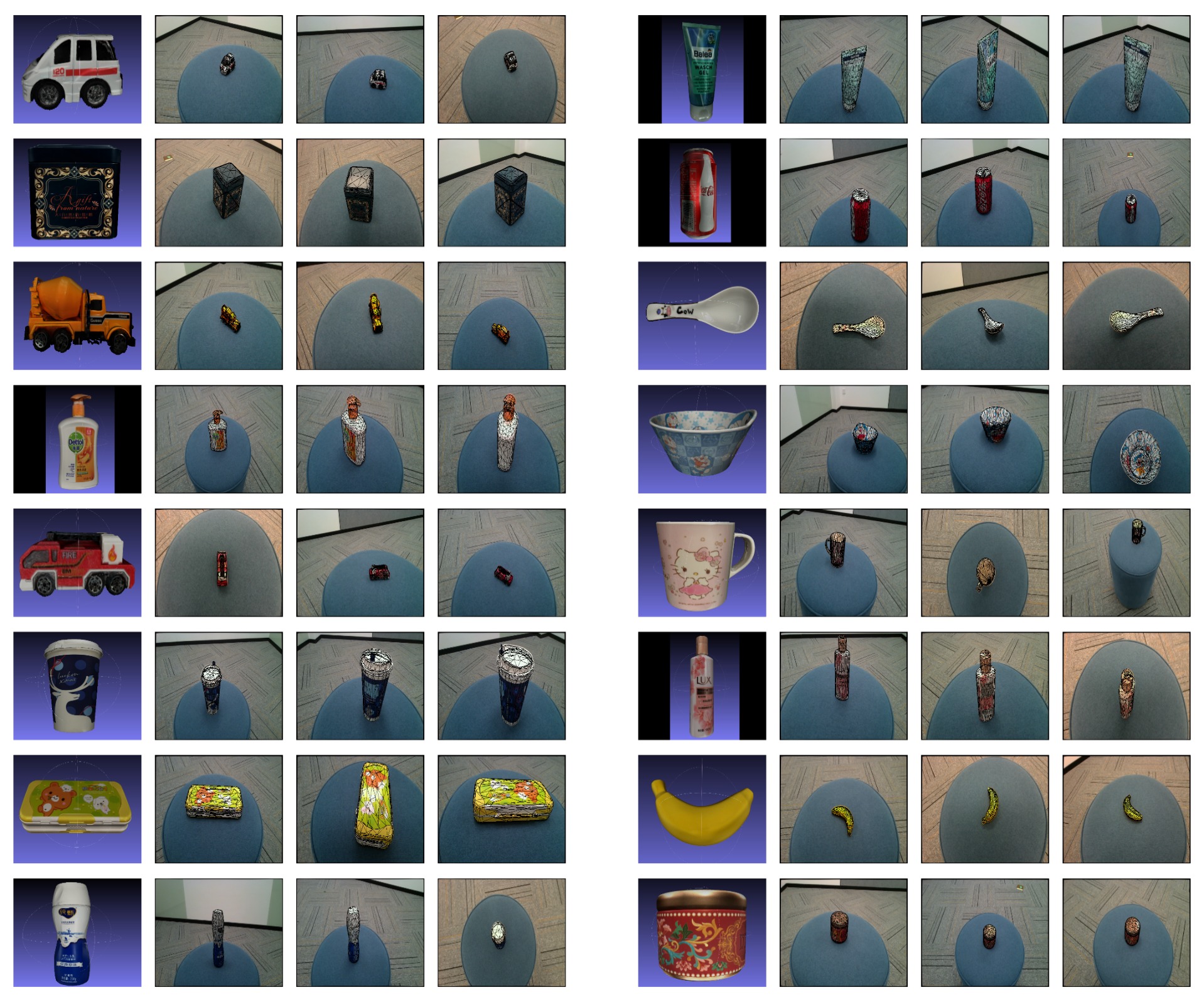}
	\vspace{-2.0em}
    \caption{\textbf{Sample data from our dataset}. From left to right, shown are visualization of textured 3D model, three sample multi-view images with wireframe object model superimposed based on annotated camera and object poses.}
	\label{fig:dataset_visualization}
	\vspace{-1.5em}
\end{figure*}

Deep learning has shown immense potential in the field of 3D vision in recent years, advancing challenging tasks
such as 3D object reconstruction, pose estimation, shape retrieval, robotic grasping etc.
But unlike for 2D tasks
~\cite{deng2009imagenet,lin2014microsoft,kuznetsova2020open}, large scale real world datasets for 3D object understanding is scarce.
Hence, to allow for further advancement of state-of-the-art in 3D object understanding we introduce our dataset
which consists of 998 high resolution, textured 3D models of everyday tabletop objects along with their 847K real world RGB-D images.
Accurate annotation of camera pose and object pose is performed for each image. 
\figref{dataset_visualization} shows some sample data from our dataset.

We primarily focus on learned multi-view 3D reconstruction due to the lack of real world datasets for the task.
3D reconstruction methods~\cite{gkioxari2019mesh,wang2018pixel2mesh,runz2020frodo,shrestha2021meshmvs,wen2019pixel2mesh++}
learn to predict 3D model of an object from its color images with known camera and object poses.
They require large amount of training examples to be able to generalize to unseen images.
While datasets like Pix3D~\cite{sun2018pix3d}, PASCAL3D+\cite{xiang2014beyond} and ObjectNet3D~\cite{xiang2016objectnet3d} provide 3D models and real world images,
they are mostly limited to a single image per model.

Existing multi-view 3D reconstruction methods~\cite{choy20163d,kar2017learning,runz2020frodo,shrestha2021meshmvs,wen2019pixel2mesh++}
rely heavily on synthetic datasets, especially ShapeNet~\cite{chang2015shapenet}, for training and evaluation.
There are a few works~\cite{lei2020pix2surf,runz2020frodo}  utilizing real world datasets~\cite{choi2016large}, but only for qualitative evaluation purpose, not for training or quantitative evaluation.
To remedy this, we present our dataset and validate its usefulness by performing training as well as qualitative/quantitative evaluation with various state-of-the-art multi-view 3D reconstruction baselines.

%

The contributions of our work are as follows:
\begin{enumerate}
    \item To the best our knowledge, our dataset is the first real world dataset that can be used for training and quantitative evaluation of learning-based multi-view 3D reconstruction algorithms.
    \item We present two novel methods for automatic/semi-automatic data annotation. We will make the annotation tools publicly available to allow future extensions to the dataset.
\end{enumerate}

\section{Related Work}

\paragraph{\bf{3D Shapes Dataset:}}
Datasets like Princeton shape benchmark \cite{shilane2004princeton}, FAUST~\cite{bogo2014faust}, ShapeNet~\cite{chang2015shapenet}
provide a large collection of 3D CAD models of diverse objects, but without associated real world RGB images.
PASCAL3D+\cite{xiang2014beyond} and ObjectNet3D~\cite{xiang2016objectnet3d}
performed rough alignment between images from existing datasets and 3D models from online shape repositories.
IKEA~\cite{Lim2013Ikea} also performed 2D-3D alignment between existing datasets but with finer alignment results on a smaller set of images and shapes (759 images and 90 shapes).
Pix3D~\cite{sun2018pix3d} extended IKEA to 10K images and 395 shapes through crowdsourcing and scanning some objects manually.
These datasets mostly have single-view images associated with the shapes. 

Datasets like~\cite{5980382,calli2015benchmarking,hodan2017t} have utilized RGB-D sensors to capture relatively small number of objects
and are mostly geared towards robot manipulation tasks rather than 3D reconstruction.
Knapitsch \etal~\cite{knapitsch2017tanks} provided a small number of large scale scenes which are suitable
for benchmarking traditional Structure-from-Motion (SfM) and Multi-view Stereo (MVS) algorithms rather than learned 3D reconstruction.

The dataset that is closest to ours is Redwood-OS~\cite{choi2016large}.
It provides RGB-D videos of 398 objects and their 3D scene reconstructions.
There are several crucial limitations that has prevented widespread adoption of this dataset for multi-view 3D reconstruction though.
Firstly, the dataset is not annotated with camera and object pose information.
While the camera pose can be obtained using Simultaneous Localization and Mapping (SLAM) or
Structure-from-Motion (SfM) techniques~\cite{cadena2016past,engel2014lsd,mur2015orb,schoenberger2016sfm,schoenberger2016mvs}, obtaining accurate object poses is relatively harder.
Also, the 3D reconstructions were performed on scene level rather than object level, making it difficult to directly use it for supervision of object reconstruction.

More recently, Objectron~\cite{ahmadyan2021objectron} and CO3D~\cite{reizenstein21co3d} have provided large scale video sequences of real world objects along with point clouds and object poses but without precise dense 3D models.
We aim to tackle the shortcomings of the existing datasets and create a dataset that can effectively serve as a real world benchmark for learning-based multi-view 3D reconstruction models.
\tableref{datasets_comparison} shows the comparison between the relevant datasets.

\begin{table*}[ht]
\begin{center}
\resizebox{\linewidth}{!}{
\begin{tabular}{|c|c|c|c|c|c|c|c|c|}
    \hline
                            & Ours      & Objectron & CO3D  & Redwood-OS    & Pix3D     & IKEA      & PASCAL3D+ & ObjectNet3D \\
    \hline
    Multi-view images       & \cmark    & \cmark    & \cmark& \cmark        & \xmark    & \xmark    & \xmark    & \xmark \\
    Dense 3D models         & \cmark    & \xmark    & \xmark& \cmark        & \cmark    & \cmark    & \cmark    & \cmark \\
    Scanned 3D models       & \cmark    & \cmark    & \cmark& \cmark        & \starmark & \xmark    & \xmark    & \xmark \\
    Object pose annotation  & \cmark    & \cmark    & \cmark& \xmark        & \cmark    & \cmark    & \starmark & \starmark \\
    Textured 3D models      & \cmark    & \xmark    & \xmark& \xmark        & \xmark    & \xmark    & \xmark    & \xmark \\
    \hline
\end{tabular}
} 
\end{center}
\vspace{-2mm}
\caption{
    Comparison between different datasets.
    Objectron and CO3D only provide point cloud models of the objects. 
    Pix3D contains a mixture of scanned and CAD 3D models.
    PASCAL3D+ and ObjectNet3D only have rough object pose annotation,
    while the annotation is not provided in Redwood-OS.
    Only our dataset provides  precisely scanned texture-mapped 3D models that are further registered to multi-view RGB images.
}
\label{table:datasets_comparison}
\end{table*}

\paragraph{\bf{3D Reconstruction:}}
The methods in \cite{gkioxari2019mesh,groueix2018papier,pan2019deep,tang2019skeleton,wang2018pixel2mesh,yao2020front2back} predict 3D models from single-view color images.
Since a single-view image can only provide a limited coverage of a target object, multi-view input is preferred in many applications.
SLAM and Structure-from-Motion methods ~\cite{cadena2016past,engel2014lsd,mur2015orb,schoenberger2016sfm,schoenberger2016mvs} are popular ways of performing 3D reconstruction
but they struggle with poorly textured and non-Lambertian surfaces and require careful input view selection.
Deep learning has emerged as a potential solution to tackle these issues.
Early works like~\cite{choy20163d,mcrecon2017,kar2017learning} used Recurrent Neural Networks (RNN) to perform multi-view 3D reconstruction.
Pixel2Mesh++~\cite{wen2019pixel2mesh++} introduced cross-view perceptual feature pooling and multi-view deformation reasoning to refine an initial shape.
MeshMVS~\cite{shrestha2021meshmvs} predicted a coarse volume from Multi-view Stereo depths first and then applied deformations on it to get a finer shape.
All of these works were trained and evaluated exclusively on synthetic datasets due to the lack of proper real world datasets.

Some recent works like DVR~\cite{niemeyer2020differentiable}, IDR~\cite{yariv2020multiview}, Neus~\cite{wang2021neus}, Geo-Neus~\cite{fu2022geo}
have focused on unsupervised 3D reconstruction with expensive per-scene optimization for each object.
These methods encode each scene into separate Multi-layer Perceptron (MLP) that implicitly represents the scene as Signed Distance Function (SDF) or Occupancy Field.
These works have obtained impressive results on small scale datasets of real world objects~\cite{jensen2014large,yao2020blendedmvs}. Our dataset can be further applied to evaluate these methods quantitatively on a much larger scale dataset.


\section{Data Acquisition}

Our data acquisition takes place in two steps.
First, a detailed and textured 3D model of an object is generated using Shining3D\textsuperscript{\tiny\textregistered} EinScan-SE 3D scanner.
The scanner uses a calibrated turntable, a 1.3 Megapixel camera and visible light sources to obtain the 3D model of an object.
Then, an Intel\textsuperscript{\tiny\textregistered} RealSense\textsuperscript{\tiny\texttrademark} LiDAR Camera L515 is used to
record a RGB-D video sequence of the object on a round ottoman chair, capturing 360{\textdegree} view around the object.
The video is recorded at 30 frames per second in HD resolution (1280$\times$720).
\figref{dataset_visualization} shows a number of 3D models and some sample color images from our dataset.

Datasets like~\cite{choi2016large,5980382} perform 3D model generation and video recording in one step by reconstructing the 3D scene captured by the images.
The quality of the 3D models generated this way depends heavily on the trajectory of the camera and requires some level of expertise for data collection.
Furthermore, these datasets use consumer grade cameras which cannot reconstruct fine details in the 3D geometry.
We therefore use specialized hardware designed for high quality 3D scanning.

Another approach is to utilize 3D CAD models from online repositories and match them with real world 2D images, which are also mostly collected online~\cite{dai2017scannet,Lim2013Ikea,xiang2016objectnet3d,xiang2014beyond}.
The downside of this approach is that it is difficult to ensure exact instance-level match between 3D models and 2D images.
According to a survey conducted by Sun \etal~\cite{sun2018pix3d}, test subjects reported that only a small fraction of the images matched the corresponding shapes in datasets~\cite{xiang2016objectnet3d,xiang2014beyond}.

\section{Data Annotation}

The most challenging aspect of creating a large scale real world dataset for object reconstruction is generating ground truth annotations.
Most learning-based 3D reconstruction methods require accurate camera poses as well as consistent object poses in the camera coordinate frame.
While it is fairly easy to obtain the camera poses, obtaining accurate object poses is more challenging.

The methods in \cite{sun2018pix3d,xiang2014beyond} perform object pose estimation by manually annotating corresponding keypoints in the 3D models and 2D images,
and then performing 2D-3D alignment with the Perspective-n-Point (PnP)~\cite{gao2003complete,lepetit2009epnp} and Levenberg-Marquardt algorithms~\cite{more1978levenberg}.
Note that these datasets mostly contain a single image for each 3D model, which makes this kind of annotation feasible. In comparison, we aim to do this for video sequences with up to 1000 images, which could be manual intensive.
Additionally, estimating object pose that is consistent over multi-view images will require keypoint matches at sub-pixel accuracy which is impossible by manual annotation.

On the other hand, the methods in \cite{dai2017scannet,xiang2016objectnet3d}  manually annotate the object pose directly by either trying to align the 3D model with the scene reconstruction~\cite{dai2017scannet} or the re-projected 3D model with 2D image\cite{xiang2016objectnet3d}.
We found these techniques to be inadequate for producing multi-view consistent object poses and therefore develop our own annotation systems.

\subsection{Notations}
We represent an object pose by $\xi \in \text{\emph{SE}(3)}$ where \emph{SE}(3) is the
3D Special Euclidean Lie group~\cite{varadarajan2013lie} of 4$\times$4 rigid body transformation matrix:
\begin{equation}
    \xi = \begin{bmatrix}
        R & \quad & t \\
        0  & \quad & 1
    \end{bmatrix}
    \label{equ:se3}
\end{equation}
\noindent
where $R$ is the 3$\times$3 rotation matrix and $t$ is the 3D translation vector.

We define object pose ${}_{\text{w}} \xi_{\text{obj}}$ as the transformation from canonical object frame (obj) to world frame (w).
Similarly, the pose of the $i^{\text{th}}$ camera ${}_{\text{w}} \xi_{\text{cam}_i}$ represents the transformation from camera to world frame.
The canonical object frame is centered at the object with z-axis pointing upwards along the gravity direction while the world frame is arbitrary (\eg pose of the first camera).

We use pinhole camera model with camera intrinsics matrix $\emph{K}$:
\begin{equation}
    \emph{K} = \begin{bmatrix}
        f_x & \quad & 0     & \quad & c_x \\
        0   & \quad & f_y   & \quad & c_y \\
        0   & \quad & 0     & \quad & 1
    \end{bmatrix}
    \label{equ:intrinsics}
\end{equation}
\noindent
where $f_x$ and $f_y$ are focal lengths and $c_x$ and $c_y$ principal points.
These parameters are provided by the camera manufacturers.

The image coordinates $\emph{p}$ of a 3D point $\emph{P}_w$ in homogeneous world coordinate can be computed as:
\begin{equation}
    \emph{p} = K
        \begin{bmatrix}
            R_i^{T} & \  & -R_i^{T} t_i
        \end{bmatrix}
        \emph{P}_w
    \label{equ:projection}
\end{equation}
\noindent
where $R_i$ and $t_i$ are the rotation and translation components of the camera pose.

The images taken from our RGB-D camera suffer from radial and tangential distortion.
But for the purpose of annotation, we undistort the images so that the pinhole camera model holds.
\\

We now present two methods for annotating our dataset depending on the texture-richness of the object being scanned:
\textbf{Texture-rich Object Annotation} and \textbf{Textureless Object Annotation}.

\subsection{Texture-rich Object Annotation}
\label{subsec:texturerich_annotation}

Since our 3D models have high-fidelity textures from our 3D scanner, we can utilize it to annotate the object pose in the recorded video sequence.
We perform joint camera and object pose estimation by matching keypoints between images and 3D model to ensure camera and object pose consistency over multiple views.
\figref{pipeline1} illustrates the annotation process.
Following are the steps involved:

\begin{figure*}[t]
	\centering
	\scriptsize
	\includegraphics[width=\textwidth]{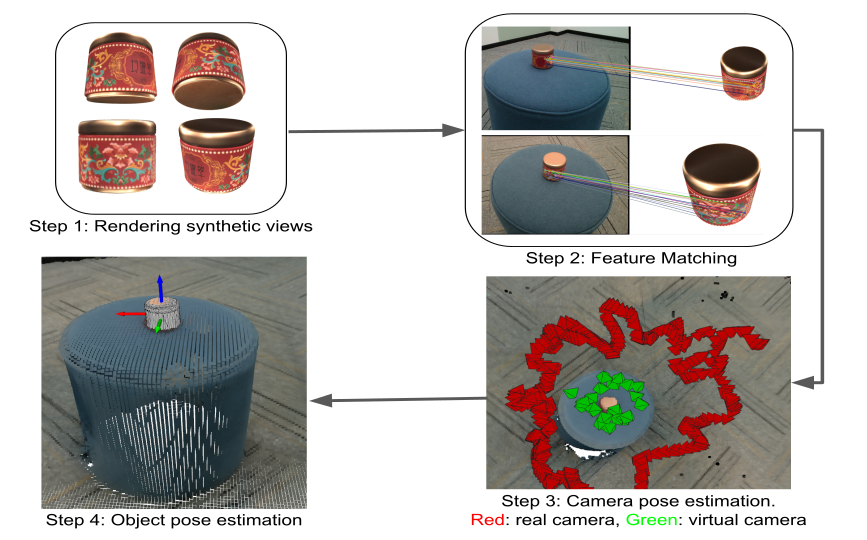}
	\vspace{-2.0em}
    \caption{
        \textbf{Texture-rich Object Annotation}.
        \emph{Step 1}: Synthetic views of the 3D model are rendered.
        \emph{Step 2}: Feature matching is performed between/across real and synthetic images.
        \emph{Step 3}: Pose of the real and virtual cameras are estimated.
        \emph{Step 4}: Object pose is estimated by 7-DOF alignment between estimated and ground truth virtual camera poses.
    }
	\label{fig:pipeline1}
	\vspace{-1.5em}
\end{figure*}

\paragraph{\bf{i. Rendering synthetic views of a 3D model:}}
Instead of directly matching keypoints between a 3D model and 2D images, we instead render synthetic views of the 3D model and perform 2D keypoint matching.
We use the physically based rendering engine, Pyrender~\cite{pyrender}, to render synthetic views.
This allows us to utilize robust keypoint matching algorithms developed for RGB images.
The virtual camera poses for rendering are randomly sampled around the object by varying the camera distance, and azimuth/elevation angles with respect to the object.
We verify the quality of each rendered image by checking if there are sufficient keypoint matches against the real images.
150 images are rendered for each object model.

\paragraph{\bf{ii. Feature matching:}}
We perform exhaustive feature matching across as well as within the real and synthetic images
using neural network based feature matching technique SuperGlue~\cite{sarlin2020superglue}.


\paragraph{\bf{iii. Camera pose estimation:}}
Given the keypoint matches, we estimate the camera poses of both the real and virtual cameras in the same world coordinate frame using the SfM tool COLMAP~\cite{schoenberger2016sfm,schoenberger2016mvs}.

\paragraph{\bf{iv. Object pose estimation:}}
Let $\{\hat{\xi}_i\ |\ i=1,...,150\}$ be the ground truth poses of the virtual cameras in object frame (we keep track of the ground truth poses during the rendering step). Let $\{\xi_{i}\ |\ i=1,...,150\}$ be the corresponding poses estimated by COLMAP in world frame.
By aligning $\{\xi_i\}$ and $\{\hat{\xi}_i\}$ 
we can estimate the object pose.
We use the Kabsch-Umeyama algorithm~\cite{umeyama1991leastsqaures} under Random Sample Consensus (RANSAC)~\cite{cantzler1981random} scheme to perform a 7-DOF (pose + scale) alignment.
Since COLMAP only uses 2D image information, its poses have arbitrary scale;
hence we perform a 7-DOF alignment instead of 6-DOF to obtain metric scale. 
After applying the Kabsch-Umeyama algorithm we get 7-DOF transformation $S$ in \emph{Sim}(3) Lie Group parameterized as:

\begin{equation}
    S = \begin{bmatrix}
        sR_s & \quad & t_s \\
        0 & \quad  & 1
    \end{bmatrix}
    \label{equ:sim3}
\end{equation}

The camera poses from COLMAP can then be transformed to metric scale pose:
\begin{equation}
    {}_{\text{w}} \xi_{\text{cam}_i} = \begin{bmatrix}
        R_s R_i & \quad & sR_s t_i + t_s \\
        0  & \quad & 1
    \end{bmatrix}
    \label{equ:sim3_transform}
\end{equation}

\noindent
where $R_i$ and $t_i$ are the rotation and translation component of the camera poses from COLMAP.

Since the ground truth virtual camera poses $\{\hat{\xi}_i | i=1,...,150\}$ are in object frame,
the transformation in \equref{sim3_transform} will lead to camera poses in object frame \ie ${}_{\text{w}} \xi_{\text{obj}} = \mathbb{I}$ where $\mathbb{I}$ is the 4$\times$4 identity matrix.

\subsection{Textureless Object Annotation}

While the pipeline outlined in \subsecref{texturerich_annotation} can accurately annotate texture-rich objects,
it will fail for textureless objects since correct feature matches among the images cannot be established.
To tackle this problem we develop another annotation system shown in~\figref{pipeline2} that can handle objects lacking good textures which consists of the following steps:

\begin{figure*}[t]
	\centering
	\scriptsize
	\includegraphics[width=\textwidth]{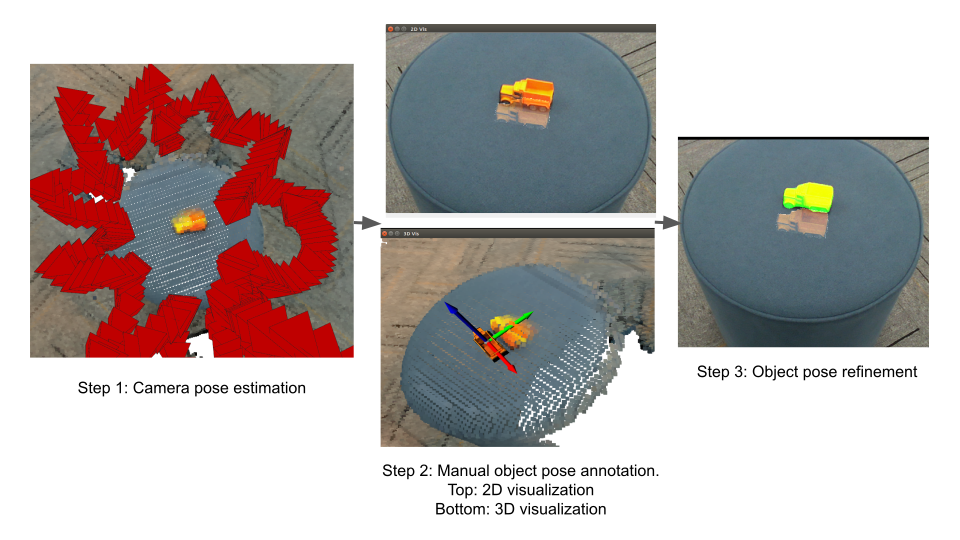}
	\vspace{-2.0em}
    \caption{
        \textbf{Textureless Object Annotation}.
        \emph{Step 1}: Camera pose annotation (+ dense scene reconstruction).
        \emph{Step 2}: Manual annotation of rough object pose where a transparent projection of the object model is superimposed over an RGB image for 2D visualization (top)
        and the 3D object is placed alongside the dense scene reconstruction for 3D visualization (bottom).
        \emph{Step 3}: Object pose is refined such that the object projection overlaps with the ground truth mask (green).
    }.
	\label{fig:pipeline2}
	\vspace{-1.5em}
\end{figure*}

\paragraph{\bf{i. Camera pose estimation:}}
Even when the object being scanned is textureless, our background has sufficient textures to allow successful camera pose estimation.
We therefore utilize the RGB-D version of ORB-SLAM2~\cite{mur2015orb} to obtain the camera poses $\{ {}_{\text{w}} \xi_{\text{cam}_i} \}$.
Since it uses depth information alongside RGB, the poses are in metric scale. 

\paragraph{\bf{ii. Manual annotation of rough object pose:}}
We create an annotation interface as shown in Step 2 of \figref{pipeline2} to estimate the rough object pose.
To facilitate the annotation, we reconstruct the 3D scene using the RGB-D images and camera poses estimated in the previous step
by employing Truncated Signed Distance Function (TSDF) fusion~\cite{zhou2013dense}. 
The object pose ${}_{\text{w}} \xi_{\text{obj}}$ is initialized to be a fixed distance in front of the first camera
and the z-axis is aligned with the principle axis of the 3D scene found using Principal Component Analysis (PCA).
An annotator can then update the 3 translation and 3 Euler angle (roll-pitch-yaw) components of the 6D object pose using keyboard to align the object model with the scene.
In addition to the 3D scene, we also show the projection of the object model over an RGB image. 
The RGB image can be changed to verify the consistency of the object pose over multiple views.

\paragraph{\bf{iii. Object pose refinement:}}
We find that obtaining accurate object pose through manual annotation is difficult,
so we refine it further by aligning the projection of the 3D object model with ground truth object masks in different images.
The ground truth object masks are obtained from Cascade Mask R-CNN~\cite{he2017mask} with a 152-layer ResNetXt backbone pretrained on ImageNet.

Let ${}_{\text{w}} \xi_{\text{obj}}$ be the rough object pose from manual annotation
and ${}_{\text{w}} \xi_{\text{cam}_i}$ be the pose of the $i^{\text{th}}$ camera.
The camera-centric object pose is represented as follows:

\begin{equation}
    \xi =  {}_{\text{cam}_i} \xi_{\text{obj}} = {(_{\text{w}} \xi_{\text{cam}_i})^{-1}} \times {}_{\text{w}} \xi_{\text{obj}}
    \label{equ:camera_centric_pose}
\end{equation}

The transformation $\xi \in \text{\emph{SE}(3)}$ is used to differentiably render~\cite{ravi2020pytorch3d}
the object model onto the image of camera $i$ to obtain the rendered object mask by applying the projection model of \equref{projection}.
Since direct optimization in the manifold space \emph{SE}(3) is not possible, we instead optimize the linearized increment of the manifold around $\xi$.
This is a common technique in SLAM and Visual Odometry~\cite{engel2014lsd,mur2015orb}.

Let $\delta \xi \in \mathfrak{se}(3)$ represent the linearized increment of $\xi$
belonging to the Lie algebra $\mathfrak{se}(3)$ corresponding to Lie Group $\emph{SE}(3)$~\cite{varadarajan2013lie}.
The updated object pose is given by:

\begin{equation}
    \xi' = \xi \times exp(\delta \xi)
    \label{equ:pose_update}
\end{equation}

\noindent
Here, $exp$ represents the exponential map that transforms $\mathfrak{se}(3)$ to $\emph{SE}(3)$.
The object pose \wrt world frame can also be updated by right multiplication of the initial pose with $exp(\delta \xi)$.
\\

We can optimize $\delta \xi$ in order to increase the overlap between the rendered mask $M$ at $\xi'$ and ground truth mask $\hat{M}$
using least-squares minimization of the mask loss:

\begin{equation}
    \mathcal{L}_{\text{mask}} = \text{mean} (\norm{M \ominus \hat{M}}_{2})
    \label{equ:mask_loss}
\end{equation}

\noindent
where $\ominus$ represents element-wise subtraction.
\\

The optimization is performed using stochastic gradient descent for each camera for 30 iterations in PyTorch~\cite{NEURIPS2019pytorch} library.
Since $\delta \xi \in \mathfrak{se}(3)$ cannot represent large changes in pose, we update the pose $\xi \leftarrow \xi'$ every 30 iterations
and relinearize $\delta \xi$ around the new $\xi$.

\section{Dataset Statistics}

We collected in total 998 objects. It typically takes about 20 minutes to scan the 3D model of an object and record a video, but about 2 hours to register the scanned 3D model to all the video frames. 
\tableref{annotation_statistics} shows the category distribution of objects in our dataset along with the method used to annotate the object (texture-rich vs textureless).
Each category in our dataset contains 39-115 objects, with average 67 objects per category.
A majority of the objects (89\%) were annotated using texture-rich pipeline which requires no user input.
\tableref{image_distribution} shows the distribution of images over the categories.
We have on average 56K images for each category.

\begin{table*}[ht]
\begin{center}
\resizebox{\linewidth}{!}{
\begin{tabular}{|c|c|c|c|c|c|c|c|c|c|c|c|c|c|c|c||c|}
    \hline
    \rows{\bf{Category}}  & \rows{Bottle} & \rows{Bowl} & \rows{Cleanser} & \rows{Cup} & Eating   & \rows{Box} & \rows{Plate} & Toy    & Toy & Toy   &  Toy       &  Toy   &  Toy    &  Toy       &  \rows{Misc}   & \rows{\bf{Total}} \\
                          &               &             &                 &            & Utensils &            &              & Animal & Car & Fruit &  Aerocraft &  Boat  &  Food   &  Figure    &                & \\
    \hline
    \bf{Texture-rich} & \rows{60} & \rows{61} & \rows{51} & \rows{39} & \rows{27}  & \rows{83} & \rows{45}  & \rows{69} & \rows{115} & \rows{12} & \rows{61} & \rows{39} & \rows{82} & \rows{95} & \rows{51} & \rows{\bf{890}} \\
    \bf{Annotation} & & & & & & & & & & & & & & & & \\
    \hline
    \bf{Textureless} & \rows{0}  & \rows{11} & \rows{0} & \rows{12} & \rows{14}  & \rows{0} & \rows{6} & \rows{32} & \rows{0} & \rows{33}  & \rows{0} & \rows{0} & \rows{0} & \rows{0} & \rows{0} & \rows{\bf{108}}  \\
    \bf{Annotation} & & & & & & & & & & & & & & & & \\
    \hline
    \bf{Total} & \bf{60} & \bf{72} & \bf{51} & \bf{54} & \bf{41} & \bf{83} & \bf{51} & \bf{101} & \bf{115} & \bf{45} & \bf{61} & \bf{39} & \bf{82} & \bf{95} & \bf{51} & \bf{998} \\
    \hline
\end{tabular}
} 
\end{center}
\caption{
    \textbf{Annotation statistics}.
}
\label{table:annotation_statistics}
\end{table*}

\begin{table*}[ht]
\begin{center}
\resizebox{\linewidth}{!}{
\begin{tabular}{|c|c|c|c|c|c|c|c|c|c|c|c|c|c|c||c|}
    \hline
    \rows{Bottle} & \rows{Bowl} & \rows{Cleanser} & \rows{Cup} & Eating   & \rows{Box} & \rows{Plate} & Toy    & Toy & Toy   &  Toy       &  Toy   &  Toy    &  Toy       &  \rows{Misc}   & \rows{\bf{Total}} \\
                  &             &                 &            & Utensils &            &              & Animal & Car & Fruit &  Aerocraft &  Boat  &  Food   &  Figure    &                & \\
    \hline
        54K & 61K & 44K & 45K & 33K & 68K & 45K & 82K & 104K & 38K & 51K & 32K & 69K & 78K & 43K & 849K \\
    \hline
\end{tabular}
} 
\end{center}
\caption{
    \textbf{Image distribution} over the categories. Number of images in each category has been rounded to nearest 1000.
}
\label{table:image_distribution}
\end{table*}

\section{Evaluation}

To verify the usefulness of our dataset, we train and evaluate state-of-the-art multi-view 3D reconstruction baselines exclusively on our dataset.
From each object, we randomly sample 100 different 3-view image tuples as the multi-view inputs.
To ensure fair evaluation and avoid overfitting we split our dataset into training, testing and validation sets in approximately 70\%-20\%-10\% ratio.
The train-test-validation split is performed such that the distribution in each object category is also 70\%-20\%-10\%.
Only the data in training set is used to fit the baseline models while validation set is used to decide when to save the model parameters during training (known as checkpointing).
All the evaluation results presented here are on the test set entirely held out during the training process.

\subsection{Experiments}
We evaluate our datasets with several recent learning-based 3D reconstruction baseline methods, including Multi-view Pixel2Mesh (MVP2M)~\cite{wen2019pixel2mesh++}, Pixel2Mesh++ (P2M++)~\cite{wen2019pixel2mesh++},
Multi-view extension of Mesh R-CNN~\cite{gkioxari2019mesh} (MV M-RCNN) provided by \cite{shrestha2021meshmvs}, MeshMVS~\cite{shrestha2021meshmvs}, DVR~\cite{niemeyer2020differentiable}, IDR~\cite{yariv2020multiview} and COLMAP~\cite{schoenberger2016sfm,schoenberger2016mvs}.
We use the `Sphere-Init' version of Mesh R-CNN and `Back-projected depth' version of MeshMVS. 

MVP2M pools multi-view image features and uses it to deform an initial ellipsoid to the desired shape.
Pixel2Mesh++ deforms the mesh predicted by MVP2M by taking the weighted sum of deformation hypothesis sampled near the MVP2M mesh vertices.
MV M-RCNN improves on MVP2M with a deeper backbone, better training recipe and higher resolution initial shape.

MeshMVS first predicts depth images using Multi-view Stereo and uses the depths to obtain a coarse shape which is deformed using similar techniques as MVP2M and MV MR-CNN.
To train the depth prediction network of MeshMVS, we use depths rendered from the 3D object models since the recorded depth can be inaccurate or altogether missing at close distances.
We also evaluate the baseline MeshMVS (RGB-D) which uses ground truth depths instead of predicted depths to obtain the coarse shape, essentially performing shape completion instead of prediction.

We also include per-scene optimized baselines DVR, IDR and COLMAP which do not require training generalizable priors with 3D supervision.
DVR and IDR perform NeRF~\cite{mildenhall2020nerf} like optimization to learn 3D models from images using implicit neural representation.
COLMAP performs Structure-from-Motion (SfM) to first generate sparse point cloud which are further densified using Patch Match Stereo algorithm~\cite{schoenberger2016mvs}.
These methods require larger number of images to produce satisfactory results, hence we use 64 input images.
Since the time required to reconstruct a scene is large for these methods, we evaluate these methods only on 30 scenes from the test set - 2 from each category.

All of the baselines require the object in the images to be segmented out of the background.
We do this by rendering the 2D image masks of 3D object models using the annotated camera/object pose.
Also, we transform the images to the size and intrinsics (\equref{intrinsics}) required by the baselines before training/testing.

\paragraph{\bf{Metrics:}}
We follow recent works~\cite{gkioxari2019mesh,shrestha2021meshmvs,wen2019pixel2mesh++} and choose F1-score (harmonic mean of precision and recall) at a thresholds $\tau=0.3$ as our evaluation metric.
Precision in this context is defined as the fraction of points in predicted model within $\tau$ distance from the ground truth points
while recall is the fraction of point in ground truth model within $\tau$ distance from the predicted points.

We also report Chamfer Distance between a predicted model $P$ and ground truth model $Q$ which measures the mean distance between the closest pairs of points $\Lambda_{P,Q} = \{(p, arg\text{ min}_q\norm{p -q}): p \in P, q \in Q\}$ in the two models:

\begin{equation}
    \mathcal{L}_{\text{chamfer}}(P, Q) = |P|^{-1} \!\!\!\!\!\!\!\!\sum_{(p, q) \in \Lambda_{P,Q}}\!\!\!\!\!\!\!\!{||p-q||^{2}} + |Q|^{-1} \!\!\!\!\!\!\!\!\sum_{(q, p) \in \Lambda_{Q,P}}\!\!\!\!\!\!\!\!{||q-p||^{2}}
    \label{equ:chamfer}
\end{equation}

%
%

We uniformly sample 10k points from predicted and ground truth meshes to evaluate these metrics.
Following~\cite{fouhey2013data,gkioxari2019mesh}, we rescale the 3D models so that the longest edge of the ground truth mesh bounding box has length 10.

\begin{figure*}[t]
	\centering
	\scriptsize
	\includegraphics[width=\textwidth]{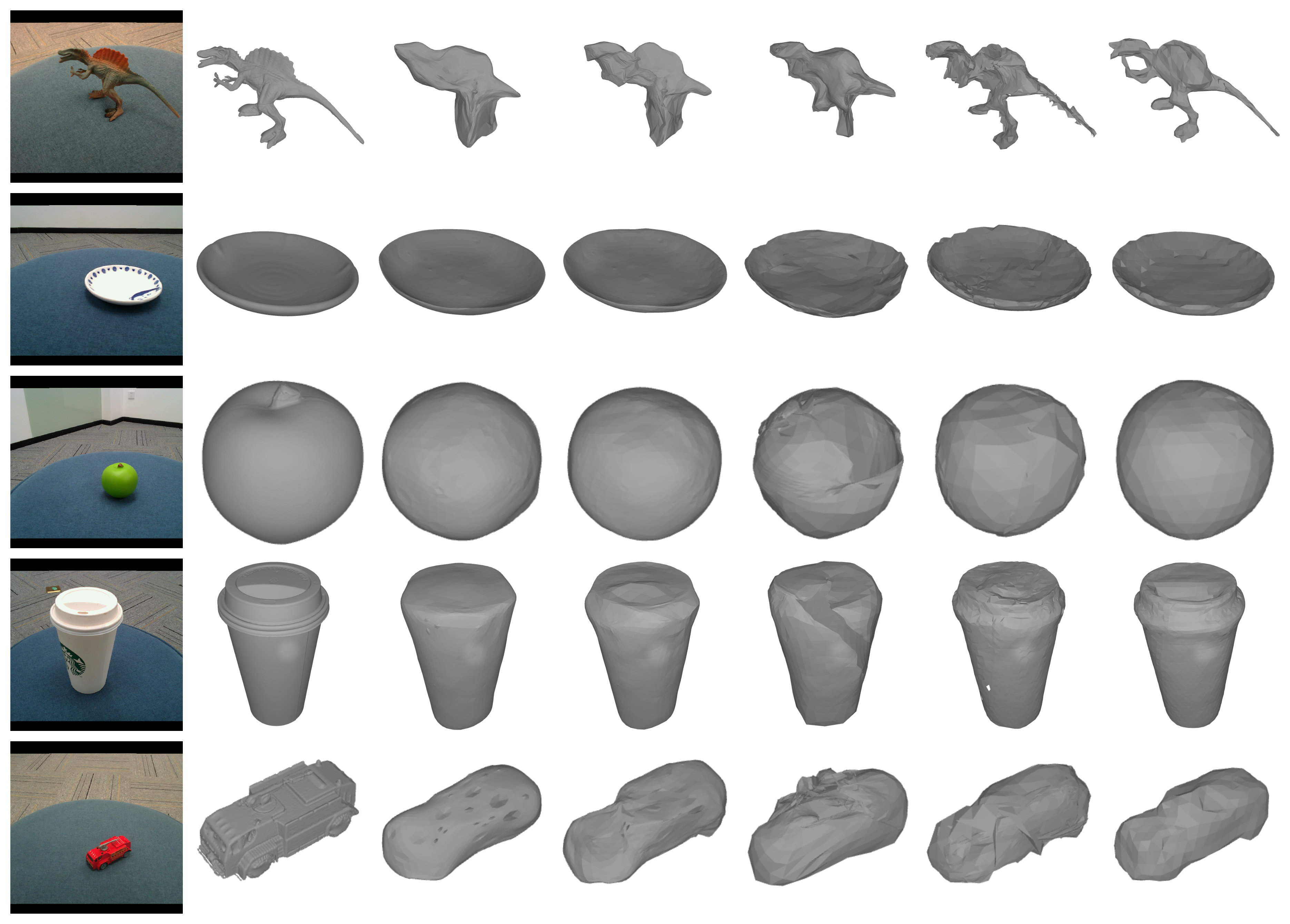}
	\vspace{-2.0em}
    \caption{\textbf{Qualitative Evaluation}. Left to right, shown are an input image, ground truth mesh, results from MVP2M, P2M++, MV M-RCNN, MeshMVS, and MeshMVS (RGB-D) respectively.}
	\label{fig:qualitative_evaluation}
	\vspace{-1.5em}
\end{figure*}

\begin{table*}[t]
\begin{center}
\resizebox{\linewidth}{!}{
\begin{tabular}{|c|ccccc|ccccc|}
    \hline
    Category&
    \multicolumn{5}{c|}{F1@0.3 $\uparrow$} &
    \multicolumn{5}{c|}{Chamfer $\downarrow$} \\
    \cline{2-11}
    & \rows{MVP2M}  & \rows{P2M++}  & MV      & \rows{MeshMVS}  & MeshMVS
    & \rows{MVP2M}  & \rows{P2M++}  & MV      & \rows{MeshMVS}  & MeshMVS \\
    &               &               & M-RCNN  &                 & (RGB-D)
    &               &               & M-RCNN  &                 & (RGB-D) \\
    \hline
    Bottle          & 74.86 & \bf{81.72} & 67.12 & 55.35 & 94.11 & 0.23 & \bf{0.19} & 1.78 & 0.53       & 0.06 \\
    Bowl            & 74.49 & \bf{80.60} & 70.94 & 59.69 & 91.88 & 0.31 & \bf{0.23} & 1.41 & 0.46       & 0.09 \\
    Box             & 61.32 & \bf{72.24} & 67.90 & 59.28 & 91.04 & 0.39 & \bf{0.29} & 0.84 & 0.48       & 0.10 \\
    Cleanser        & 68.69 & \bf{80.90} & 72.37 & 64.01 & 95.97 & 0.30 & \bf{0.19} & 0.96 & 0.39       & 0.04 \\
    Cup             & 65.31 & \bf{73.67} & 66.00 & 52.41 & 87.40 & 0.39 & \bf{0.30} & 1.59 & 0.64       & 0.12 \\
    Eating Utensils & 74.34 & \bf{88.40} & 81.67 & 72.74 & 96.08 & 0.24 & \bf{0.12} & 0.84 & 0.26       & 0.04 \\
    Plate           & 76.13 & \bf{81.65} & 66.69 & 72.03 & 82.70 & 0.34 & \bf{0.26} & 3.42 & 0.27       & 0.16 \\
    Toy Boat        & 55.77 & \bf{65.57} & 57.95 & 59.58 & 88.19 & 0.46 & \bf{0.37} & 6.54 & 5.22       & 5.79 \\
    Toy Aerocraft   & 52.85 & \bf{65.31} & 56.91 & 64.94 & 91.13 & 0.73 & 0.55      & 2.91 & \bf{0.34}  & 0.08 \\
    Toy Animals     & 49.46 & \bf{68.12} & 64.07 & 59.89 & 93.53 & 0.81 & 0.51      & 0.94 & \bf{0.42}  & 0.06 \\
    Toy Food        & 60.92 & \bf{71.24} & 61.08 & 49.12 & 90.16 & 0.35 & \bf{0.25} & 2.78 & 0.59       & 0.10 \\
    Toy Fruit       & 56.78 & \bf{72.85} & 59.84 & 41.28 & 88.86 & 0.55 & \bf{0.37} & 3.77 & 0.87       & 0.11 \\
    Miscellaneous   & 56.54 & \bf{69.40} & 66.34 & 61.57 & 91.40 & 0.63 & \bf{0.46} & 0.96 & 0.47       & 0.11 \\
    Toy Car         & 59.32 & \bf{71.33} & 63.90 & 57.65 & 88.28 & 0.37 & \bf{0.25} & 1.47 & 0.47       & 0.10 \\
    Toy Figure      & 50.42 & \bf{68.32} & 57.23 & 58.63 & 91.39 & 0.70 & \bf{0.46} & 5.12 & 0.52       & 0.08 \\
    \hline
    All             & 61.25 & \bf{73.30} & 64.77 & 58.79 & 90.85 & 0.47 & \bf{0.33} & 2.24 & 0.65       & 0.29 \\
    \hline

\end{tabular}}
\end{center}
\caption{
    \textbf{Quantitative comparison} of state-of-the-art learning-based multi-view 3D reconstruction methods on our dataset.
    We report F1-score and Chamfer Distance on each semantic category as well as over all categories.
    The baseline MeshMVS (RGB-D) is not considered for highlighting the best performance since it uses ground truth depth as additional input.
}
\label{table:learned_comparison}
\end{table*}

\begin{table*}[h]
\begin{center}
\begin{tabular}{|c|c@{\hskip 0.2in}cc|c@{\hskip 0.2in}cc|}
    \hline
    Category&
    \multicolumn{3}{c|}{F1@0.3 $\uparrow$} &
    \multicolumn{3}{c|}{Chamfer $\downarrow$} \\
    \cline{2-7}
    & DVR & IDR & COLMAP & DVR & IDR & COLMAP \\
    \hline
    Bottle          & \bf{92.95} & 91.90      & 25.80       & 0.11      & \bf{0.09} & 1.36 \\
    Bowl            & 69.20      & \bf{83.50} & 14.72       & \bf{0.60} & 0.62      & 1.81 \\
    Box             & 75.97      & \bf{78.95} & 34.64       & 0.70      & \bf{0.22} & 1.66 \\
    Cleanser        & 86.61      & \bf{98.32} & 45.54       & 0.09      & \bf{0.02} & 1.38 \\
    Cup             & 70.77      & \bf{78.69} & 14.86       & \bf{0.64} & 0.68      & 3.23 \\
    Eating utensils & 87.93      & \bf{95.12} & 41.34       & 0.08      & \bf{0.04} & 1.10 \\
    Plate           & 62.46      & 75.01      & \bf{76.98}  & 0.50      & 0.45      & \bf{0.14} \\
    Toy Boat        & 86.71      & \bf{99.87} & 45.66       & 0.08      & \bf{0.01} & 0.37 \\
    Toy Aerocraft   & 78.48      & \bf{99.07} & 86.37       & 0.24      & \bf{0.02} & 0.09 \\
    Toy Animals     & 85.26      & \bf{90.61} & 53.31       & 0.19      & \bf{0.11} & 0.67 \\
    Toy Food        & 86.06      & \bf{89.59} & 34.74       & 0.12      & \bf{0.08} & 0.96 \\
    Toy Fruit       & 63.27      & \bf{90.25} & 0.55        & 0.75      & \bf{0.09} & 40.56 \\
    Miscellaneous   & 89.21      & \bf{89.31} & 45.96       & 0.10      & \bf{0.09} & 0.66 \\
    Toy Car         & 76.25      & \bf{94.26} & 27.24       & 0.24      & \bf{0.06} & 1.32 \\
    Toy Figure      & 82.25      & \bf{97.60} & 49.39       & 0.16      & \bf{0.03} & 0.32 \\
    \hline
    All             & 79.56      & \bf{89.54} & 39.81       & 0.30      & \bf{0.17} & 3.71 \\
    \hline
\end{tabular}
\end{center}
\caption{
    \textbf{Quantitative comparison} of state-of-the-art NeRF-based 3D reconstruction methods along with COLMAP on our dataset.
    We report F1-score and Chamfer Distance on each semantic category as well as over all categories.
}
\label{table:unsupervised_comparison}
\end{table*}

\paragraph{\bf{Results:}}
The quantitative comparison results of different  learning-based 3D reconstruction baselines on our dataset are presented in~\tableref{learned_comparison}.
Note that both training and testing set contain objects from all categories, but test F1-score on individual categories as well as over all categories are reported here.
\figref{qualitative_evaluation} visualizes the shapes generated by different methods for qualitative evaluation.

We can see that overall Pixel2Mesh++ performs the best (barring MeshMVS RGB-D).
This is contrary to the results on ShapeNet reported in~\cite{shrestha2021meshmvs} where MeshMVS performs the best.
This can be attributed to the high depth prediction error of MeshMVS (average depth error is $\sim$6\% of the total depth range).
When predicted depth is replaced with ground truth depth, we indeed see a significant improvement in the performance of MeshMVS
indicating that depth prediction is the main bottleneck in its performance.

\vspace{-2.0em}
\begin{wrapfigure}{r}{0.5\textwidth}
    \begin{center}
        \includegraphics[width=0.5\textwidth]{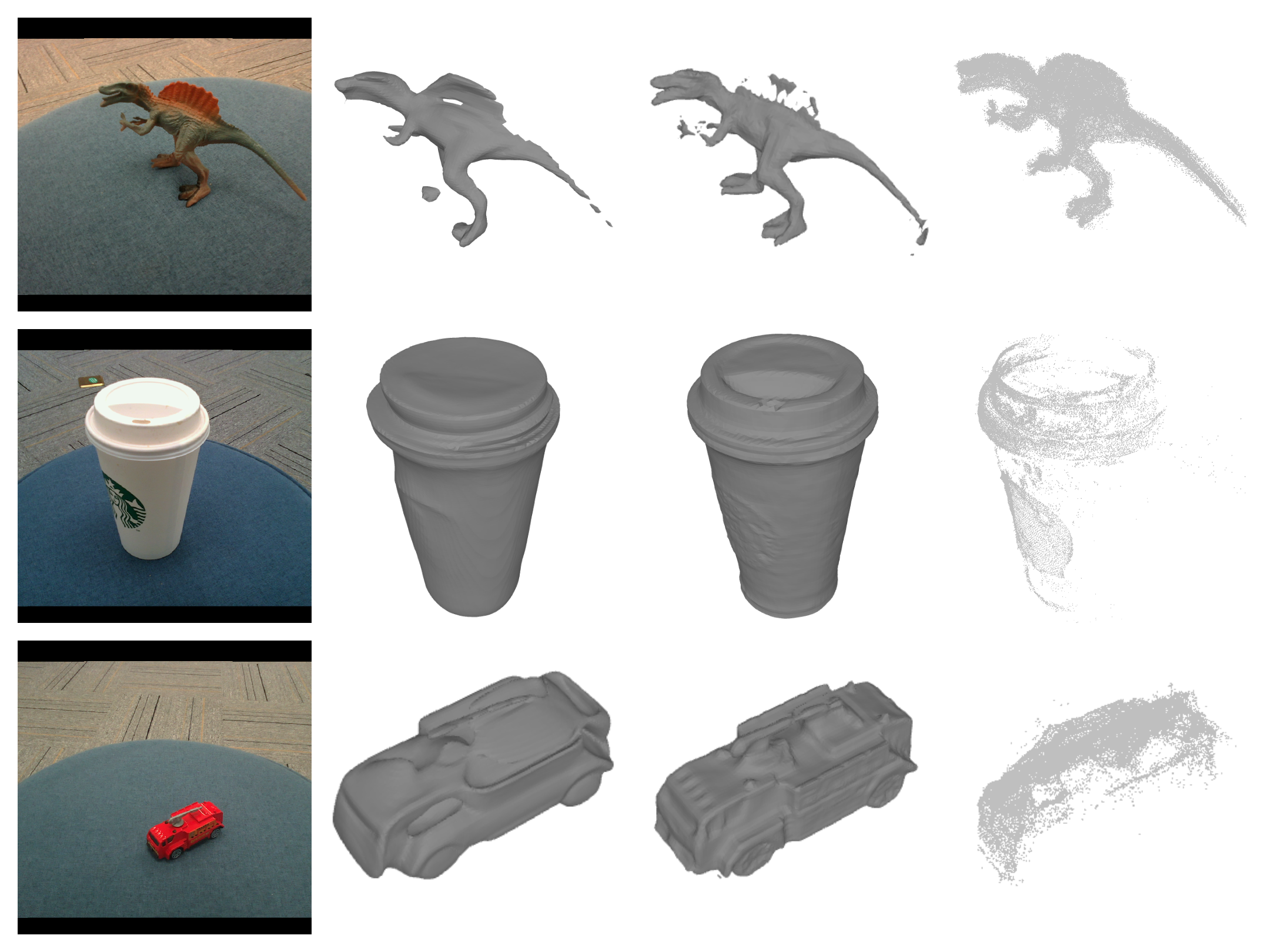}
    \end{center}
	\vspace{-1.5em}
    \caption{
        \textbf{Qualitative evaluation}. Left to right: Input image, results from DVR, IDR, COLMAP.
    }
	\label{fig:unsupervised_evaluation}
	\vspace{-1.5em}
\end{wrapfigure}

\tableref{unsupervised_comparison} shows the quantitative comparison between different unsupervised, per-scene optimized baselines.
Here, IDR outperforms the other two baselines which is in line with the results presented in~\cite{yariv2020multiview} on the DTU dataset.
COLMAP performs worse than the rest because the textures on most of the objects are insufficient for dense reconstruction using Patch Match stereo leading to sparse and noisy results (\figref{unsupervised_evaluation}).


\paragraph{\bf{Single category training:}}
We compare the difference in the performance when each category is trained and evaluated separately. 
In this case, there will be a different set of model parameters for each category.
For these experiments we sample 200 different 3-view images as inputs from each scene.
\tableref{single_category_training} shows the results for MV M-RCNN baseline when each category is trained separately versus when all are trained together.
We see that the performance is generally better when using all categories, showing that 3D reconstruction models can learn to generalize over multiple categories in our dataset.
\begin{table*}[ht]
\begin{center}
\begin{tabular}{|c|c@{\hskip 0.3in}c|c@{\hskip 0.3in}c|}
    \hline
    Category &
    \multicolumn{2}{c|}{F1@0.3 $\uparrow$} &
    \multicolumn{2}{c|}{Chamfer $\downarrow$} \\
    \cline{2-5}
    & All & Single & All & Single \\
    \hline
    Bottle          & \bf{67.12} & 64.02        & \bf{1.78}     & 3.33  \\
    Bowl            & \bf{70.94} & 53.41        & \bf{1.41}     & 27.37 \\
    Box             & \bf{67.90} & 65.81        & \bf{0.84}     & 1.85  \\
    Cleanser        & 72.37      & \bf{73.00}   & \bf{0.96}     & 1.21  \\
    Cup             & \bf{66.00} & 61.95        & \bf{1.59}     & 1.61  \\
    Eating utensils & \bf{81.67} & 77.06        & 0.84          & \bf{0.80}  \\
    Plate           & \bf{66.69} & 62.15        & \bf{3.42}     & 50.72 \\
    Toy Boat        & \bf{57.95} & 55.39        & \bf{6.54}     & 10.21 \\
    Toy Aerocraft   & \bf{56.91} & 43.51        & \bf{2.91}     & 5.16  \\
    Toy Animals     & \bf{64.07} & 62.32        & \bf{0.94}     & 1.16 \\
    Toy Food        & 61.08      & \bf{62.02}   & \bf{2.78}     & 4.92 \\
    Toy Fruit       & \bf{59.84} & 20.54        & \bf{3.77}     & 67.82 \\
    Miscellaneous   & \bf{66.34} & 44.29        & \bf{0.96}     & 3.71 \\
    Toy Car         & 63.90      & \bf{65.18}   & 1.47          & \bf{1.26} \\
    Toy Figure      & \bf{57.23} & 50.81        & \bf{5.12}     & 9.0 \\
    \hline
    Mean            & \bf{64.77} & 58.15        & \bf{2.24}     & 10.57 \\
    \hline
\end{tabular}
\end{center}
\caption{
    \textbf{Single Vs All Category Training} evaluation on MV~M-RCNN baseline.
}
\label{table:single_category_training}
\end{table*}


\section{Discussion}

The results presented in Tables~\ref{table:learned_comparison} and~\ref{table:single_category_training}
as well as the qualitative evaluation of \figref{qualitative_evaluation} show that the problem of generalizable multi-view 3D reconstruction is far from solved.
While works like Pixel2Mesh++, Mesh R-CNN and MeshMVS have offered promising avenues for advancement of the state-of-the-art, more research is still needed in this direction.
\tableref{unsupervised_comparison} and \figref{unsupervised_evaluation} shows the limitations of traditional 3D reconstruction methods like COLMAP. While more recent NeRF-based methods like DVR and IDR generates high quality reconstruction, their running time is at the order of 10 hours in general and requires a larger number of input images (64 in our case).
We hope that our dataset can serve as a challenging benchmark for these problems; aiding and inspiring future work in 3D shape generation.

\section{Conclusion}

We present a large scale dataset of 3D models and their real world multi-view images.
Two methods were developed for annotation of the dataset which can provide high accuracy camera and object poses.
Experiments show that our dataset can be used for training and evaluating multi-view 3D reconstruction methods,
something that has been lacking in existing real world datasets.

\clearpage
%
%
\bibliographystyle{splncs04}
\bibliography{egbib}
\end{document}